\documentclass[10pt,twocolumn,letterpaper]{article}

\usepackage{cvpr}
\usepackage{times}
\usepackage{epsfig}
\usepackage{graphicx}
\usepackage{amsmath}
\usepackage{amssymb}
\usepackage{multirow}
\usepackage{listings}
\usepackage{wasysym}
\newcommand{\mypara}{\vspace*{-3mm}\paragraph}

\usepackage[pagebackref=true,breaklinks=true,letterpaper=true,colorlinks,bookmarks=false]{hyperref}

\cvprfinalcopy % *** Uncomment this line for the final submission

\def\cvprPaperID{2212} % *** Enter the CVPR Paper ID here

\ifcvprfinal\fi

\begin{document}

%%%%%%%%% TITLE
\title{Physically-Based Rendering for Indoor Scene Understanding\\Using Convolutional Neural Networks} 

\vspace{-2mm}
\author{Yinda Zhang$^{\dagger*}$~~~Shuran Song$^\dagger$\thanks{∗ indicates equal contributions.}~~~Ersin Yumer$^\ddagger$~~~Manolis Savva$^\dagger$\\ Joon-Young Lee$^\ddagger$~~~Hailin Jin$^\ddagger$~~~Thomas Funkhouser$^\dagger$ \\ \\ $^\dagger$Princeton University \quad\quad $^{\ddagger}$Adobe Research }
\vspace{-5mm}

\maketitle
\thispagestyle{empty}

%%%%%%%%% ABSTRACT
\begin{abstract}
Indoor scene understanding is central to applications such as robot navigation and human companion assistance. Over the last years, data-driven deep neural networks have outperformed many traditional approaches thanks to their representation learning capabilities. One of the bottlenecks in training for better representations is the amount of available per-pixel ground truth data that is required for core scene understanding tasks such as semantic segmentation, normal prediction, and object boundary detection.
To address this problem, a number of works proposed using synthetic data. However, a systematic study of how such synthetic data is generated is missing. 
In this work, we introduce a large-scale synthetic dataset with 500K physically-based rendered images from 45K realistic 3D indoor scenes. We study the effects of rendering methods and scene lighting on training for three computer vision tasks: surface normal prediction, semantic segmentation, and object boundary detection. This study provides insights into the best practices for training with synthetic data (more realistic rendering is worth it) and shows that pretraining with our new synthetic dataset can improve results beyond the current state of the art on all three tasks. 
\end{abstract}

%%%%%%%%% BODY TEXT
\vspace{-4mm}
\section{Introduction}

\begin{figure}
\centering
 \includegraphics[width=1\columnwidth]{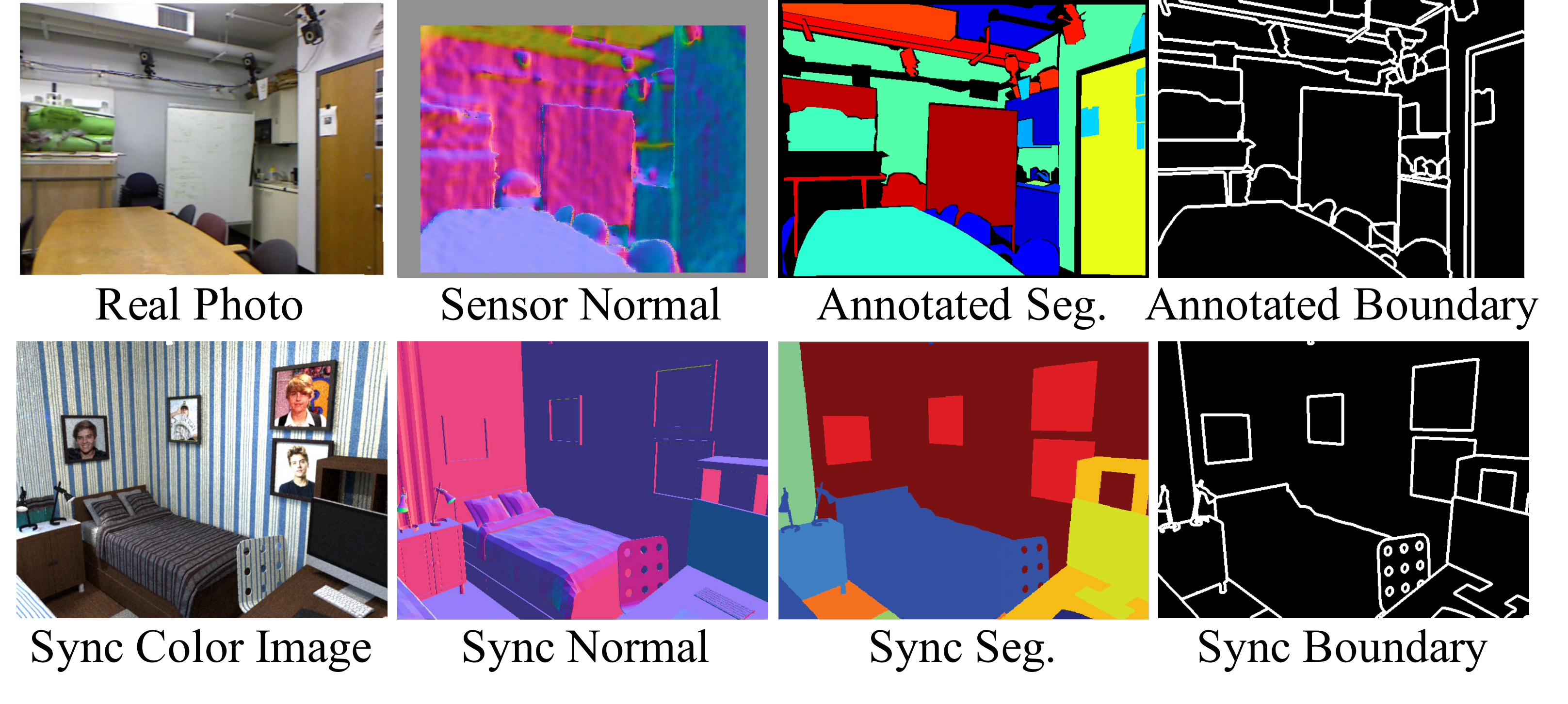}
\caption{\label{fig:motivation} Real data (top) vs. synthetic data (bottom). For the real data, note the noise in normal map and the diminishing accuracy at object boundaries in the semantic labels. }
\vspace{-2mm}
\end{figure}

% indoor scene understanding is important 
Indoor scene understanding is crucial to many applications including but not limited to robotic agent path planning, assistive human companions, and monitoring systems.
% goal statement
One of the most promising approaches to tackle these issues is using a data-driven method, where the representation is learned from large amount of data.
% real world data challenges
However, real world data is very limited for most of these tasks, such as the widely used indoor RGBD dataset for normal prediction introduced by Silberman \etal~\cite{silberman2012indoor}, which contains merely 1449 images. Such datasets are not trivial to collect due to various requirements such as depth sensing technology~\cite{silberman2012indoor,SUNRGBD} and excessive human effort for semantic segmentation~\cite{MSCOCO,everingham2010pascal}. Moreover, current datasets lack pixel level accuracy due to sensor noise or labeling error (Fig.~\ref{fig:motivation}).

% synthetic data has been coined
This has recently led to utilizing synthetic data in the form of 2D render pairs (RGB image and per-pixel label map) from digital 3D models~\cite{aubry2015understanding,dosovitskiy2015flownet,handa2015scenenet,zhang2016deep,su2015render,HowUseful}. 
% open problems with real world data
However, there are two major problems that have not been addressed: (1) studies of how indoor scene context affect training have not been possible due to the lack of large scene datasets, so training is performed mostly on repositories with independent 3D objects~\cite{chang2015shapenet}; and (2) systematic studies have not been done on how such data should be rendered; unrealistic rendering methods often are used in the interest of efficiency. 

% objects have not been treated in their natural context at the time of data generation, except for the recent SYNTHIA~\cite{ros2016synthia} dataset, which provides context only for outdoor scenes, 

\begin{figure*}
\vspace{-4mm}
\centering
\includegraphics[width=1\textwidth]{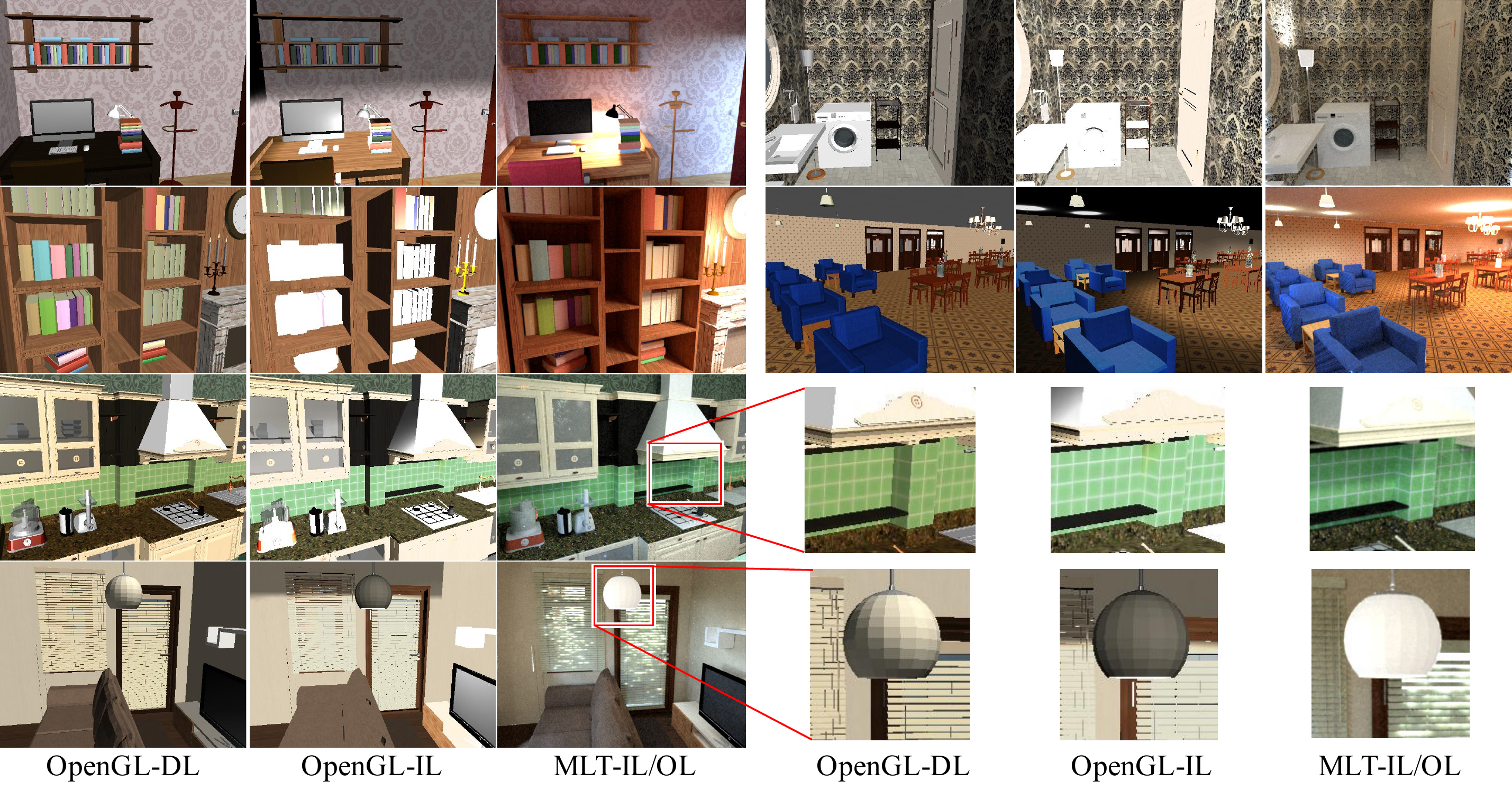}
\vspace{-5mm}
\caption{\label{fig:rendering} Render output examples with \textsc{OpenGL-DL}, \textsc{OpenGL-IL}, and \textsc{MLT-IL/OL}. The physically based rendering with proper illumination provides the best rendering quality with soft shadow and realistic material, highlighted in the zoomed in view. First two rows show four typical examples in our dataset, last two rows show two examples with zoomed in views.}
\vspace{-4mm}
\end{figure*}

% how do we address these open problems
To address these problems, we introduce a large scale (500K images) synthetic dataset that is created from 45K 3D houses designed by humans \cite{SSCNet}. Using such realistic indoor 3D environments enable us to create 2D images for training in realistic context settings where support constructs (e.g. such as walls, ceilings, windows) as well as light sources exist together with common household objects. Since we have access to the source 3D models, we can generate dense per-pixel training data for all tasks, virtually with no cost. 

Complete control over the 3D scenes enables us to systematically manipulate both outdoor and indoor lighting, sample as many camera viewpoints as required, use the shapes in-context or out-of-context, and render with either simple shading methods, or physically based based rendering. For three indoor scene understanding tasks, namely normal prediction, semantic segmentation, and object edge detection, we study how different lighting conditions, rendering methods, and object context effects performance.

% Quantitative achievements/facts of the paper
We use our data to train deep convolutional neural networks for per-pixel prediction of semantic segmentation, normal prediction, and object boundary prediction, followed by finetuning on real data. Our experiments show that for all three indoor scene understanding tasks, we improve over the state of the art performance. We also demonstrate that physically based rendering with realistic lighting and soft shadows (which is not possible without context) is superior to other rendering methods.

% Concrete contributions
In summary, our main contributions are as follows:
\vspace{-3mm}
\begin{itemize}
\setlength{\topsep}{0pt}
\setlength{\parsep}{0pt}
\setlength{\parskip}{0pt}
\setlength{\itemsep}{2pt}
	\item{We introduce a dataset with 500K synthetic image instances where each instance consists of three image renders with varying render quality, per-pixel accurate normal map, semantic labels and object boundaries. The dataset will be released.}
	\item{We demonstrate how different rendering methods effect normal, segmentation, and edge prediction tasks. We study the effect of object context, lighting and rendering methodology on performance.}
	\item{We provide pretrained networks that achieve the state of the art on all of the three indoor scene understanding tasks after fine-tuning. }
	
\end{itemize}

%------------------------------------------------------------------------

\section{Background}
Using synthetic data to increase the data density and diversity for deep neural network training has shown promising results. To date, synthetic data have been utilized to generate training data for predicting object pose~\cite{su2015render,HowUseful,gupta2015aligning}, optical flow~\cite{dosovitskiy2015flownet}, semantic segmentation~\cite{handa:etal:ICRA2014,handa2015scenenet,zhang2016deep,richter2016playing}, and investigating object features~\cite{aubry2015understanding,EvalVirtualWorld}. 

Su \etal~\cite{su2015render} used individual objects rendered in front of arbitrary backgrounds with prescribed angles relative to the camera to generate data for learning to predict object pose. Similarly, Dosovitskiy \etal~\cite{dosovitskiy2015flownet} used individual objects rendered with  arbitrary motion to generate synthetic motion data for learning to predict optical flow. Both works used unrealistic OpenGL rendering with fixed lights, where physically based effects such as shadows, reflections were not taken into account. Movshovitz \etal~\cite{HowUseful} used environment map lighting and showed that it benefits pose estimation. However, since individual objects are rendered in front of arbitrary 2D backgrounds, the data generated for these approaches lack correct 3D illumination effects due to their surroundings such as shadows and reflections from nearby objects with different materials. Moreover, they also lack realistic context for the object under consideration.

Handa \etal~\cite{handa:etal:ICRA2014,handa2015scenenet} introduced a laboriously created 3D scene dataset and demonstrated the usage on semantic segmentation training. However, their data consisted rooms on the order of tens, which has significantly limited variation in context compared to our dataset with 45K realistic house layouts. Moreover, their dataset has no RGB images due to lack of colors and surface materials in their scene descriptions, hence they were only able to generate depth channels. Zhang \etal~\cite{zhang2016deep} proposed to replace objects in depth images with 3D models from ShapeNet~\cite{chang2015shapenet}. However, there is no guarantee whether replacements will be oriented correctly with respect to surrounding objects or be stylistically in context. In contrast, we take advantage of a large repository of indoor scenes created by human, which guarantees the data diversity, quality, and context relevance. 

Xiang \etal~\cite{xiang2016objectnet3d} introduced a 3D object-2D image database, where 3D objects are manually aligned to 2D images. The image provides context, 
however the 3D data contains only the object without room structures, it is not possible to extract per-pixel ground truth for the full scene. The dataset is also limited with the number of images provided (90K). In contrast, we can provide as many (rendered image, per-pixel ground truth) pairs as one wants.

Recently, Richter \etal~\cite{richter2016playing} demonstrated collecting synthetic data from realistic game engine by intercepting the communication between game and the graphics hardware. They showed that the data collected can be used for semantic segmentation task. Their method ensures as much context as there is in the game (Although it is limited to only outdoor context, similar to the SYNTHIA~\cite{ros2016synthia} dataset). However they largely reduced the human labor in annotation by tracing geometric entities across frames, the ground truth (i.e. per-pixel semantic label) collection process is not completely automated and error prone due to the human interaction: even though they track geometry through frames and propagate most of the labels, a person needs to label new objects emerging in the recorded synthetic video. Moreover, it is not trivial to alter camera view, light positions and intensity, or rendering method due to lack of access to low level constructs in the scene. On the other hand, our data and label generation process is automated, and we have full control over how the scene is lit and rendered.

\begin{figure}
\vspace{-3mm}
\centering
 \includegraphics[width=1\columnwidth]{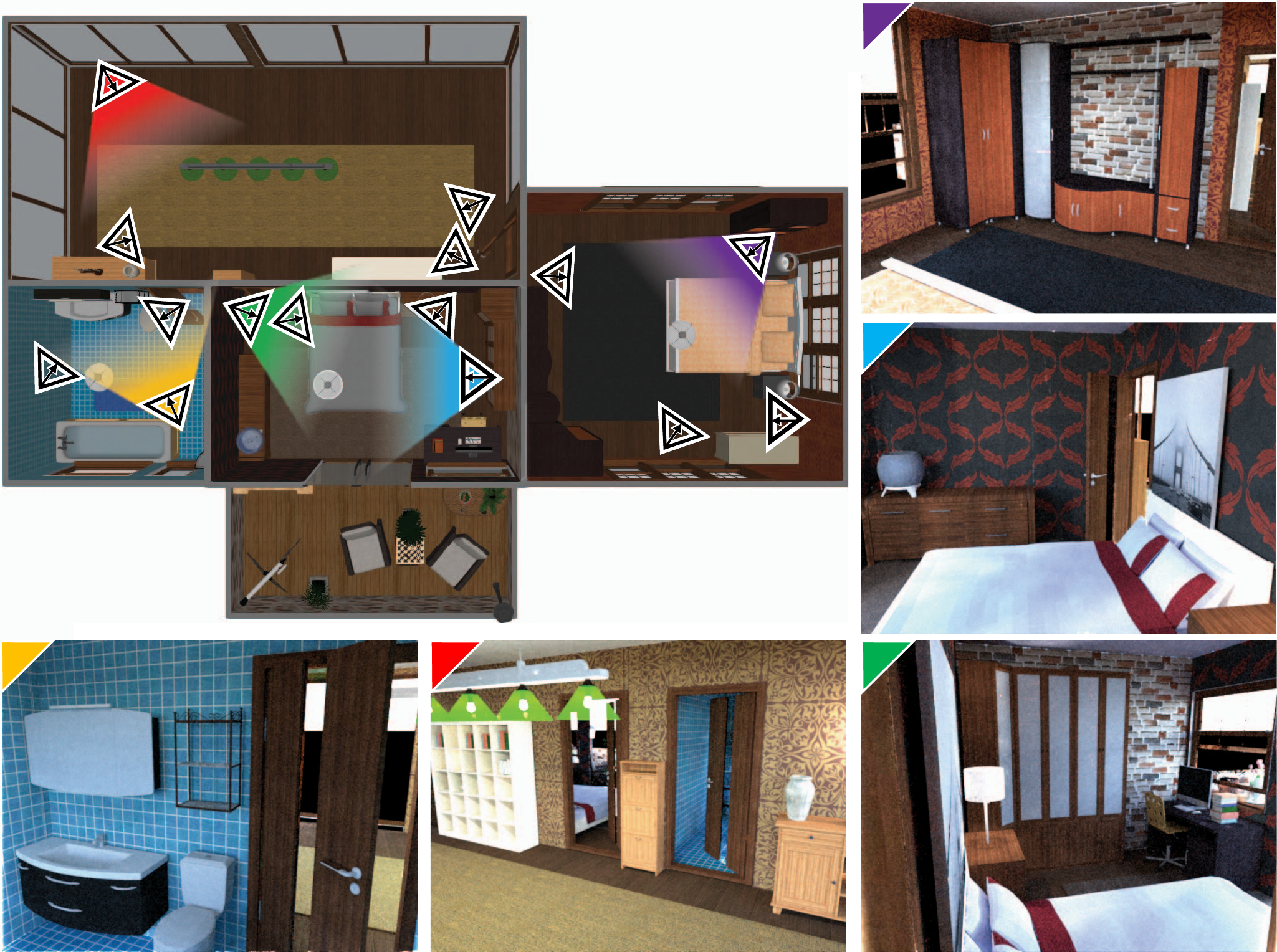}
\caption{\label{fig:camerasamples} Typical camera samples in our dataset, and corresponding images rendered from these viewpoints.}
\vspace{-4mm}
\end{figure}

%------------------------------------------------------------------------

\section{Data}\label{sec:data}

We modify the 3D scene models from the SUNCG dataset \cite{SSCNet} to generate synthetic data.
%We use a collection of 3D scene models downloaded from SUNCG dataset \cite{SSCNet} [xxxxxcarexxxx]. 
In SUNCG, there are 45,622 scenes with over 5M instances of 2644 unique objects in 84 object categories. 
The object models provide surface materials, including reflectance, texture, and transparency, which are used to obtain photo-realistic renderings. One of the important aspects of this dataset is the fact that the indoor layouts, furniture/object alignment, and surface materials are designed by people to replicate existing settings.
%, or to plan for new upgrades for current homes. 
However, these raw 3D models lack sufficiently accurate geometry (e.g. solid walls) and materials (e.g. emissive surfaces for lighting) for physically based rendering.
We fix these problems, and release the accurate full 3D scene models ready for rendering on our \href{http://pbrs.cs.princeton.edu}{project webpage}.
%We will release our version of this data, together with minor improvements we have made (such as adding indoor light source labels, as explained later in this section), as well as our camera settings in order to facilitate repeatability. 

\vspace{-1mm}
\subsection{Camera Sampling} 
\label{camerasampling}
\vspace{-1mm}
For each scene, we select a set of cameras with a process that seeks a diverse set of views seeing many objects in context. Our process starts by selecting the ``best'' camera for each of six horizontal view direction sectors in every room.  For each of the six views, we sample a dense set of cameras on a 2D grid with 0.25 resolution, choosing a random viewpoint within each grid cell, a random horizontal view direction within the 60 degree sector, a random height 1.5-1.6m above the floor, and a downward tilt angle of 11 degrees, while excluding viewpoints within 10cm of any obstacle to simulate typical human viewing conditions. For each of these cameras, we render an item buffer and count the number of pixels covered by each visible ``object'' in the image (everything except wall, ceiling, and floor).  For each view direction in each room, we select the view with the highest pixel coverage, as long it has at least three different visible objects each covering at least $1\%$ of the pixels.  This process yields 6N candidate cameras for N rooms. Figure~\ref{fig:camerasamples} shows the cameras sampled from an example house. 

\vspace{-1mm}
\subsection{Image Rendering}
\label{renderingmethods}
\vspace{-1mm}
We render images from these selected cameras using four combinations of rendering algorithms and lighting conditions, ranging from fast/unrealistic rendering with directional lights using the OpenGL pipeline to physically-based rendering with local lights using Mitsuba.

\mypara{OpenGL with Directional Lights (\textsc{OpenGL-DL}).}  Our first method renders images with the OpenGL pipeline. The scene is illuminated with three lights: a single directional headlight pointing along the camera view direction and two directional lights pointing in nearly opposite diagonal directions with respect to the scene.  No local illumination, shadows, or indirect illumination is included.

\mypara{OpenGL with Indoor Lights (\textsc{OpenGL-IL}).}  Our second method also uses the OpenGL pipeline.  However, the scene is augmented with local lights approximating the emission of indoor lighting appliances.   For each object emitting light, we create a set of OpenGL point lights and spot lights approximating its emission patterns.  We then render the scene with these lights enabled (choosing the best 8 lights sources for each object based on illumination intensity), and no shadows or indirect illumination is included.

\begin{figure}
\vspace{-3mm}
\centering
\includegraphics[width=1\columnwidth]{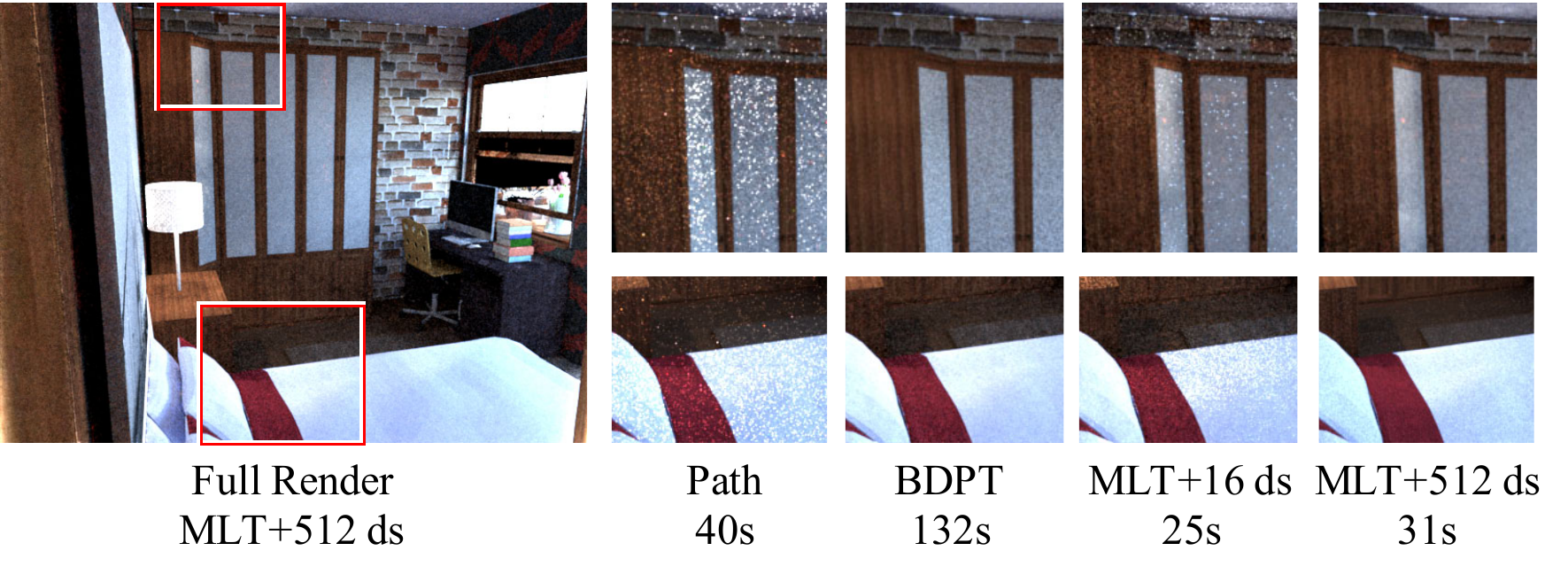}
\vspace{-4mm}
\caption{\label{fig:rendertech} Quality and running time of different rendering techniques. Path tracing does not converge well and introduces white dot artifacts. Bidirectional path tracing works well but is very slow. Metropolis Light Transport (MLT) with low sampler rate for direct illumination still occasionally introduces white dot artifacts. We take MLT with high sampler rate for direct illumination.}
\vspace{-2mm}
\end{figure}

\mypara{Physically Based Rendering with Outdoor Lights (\textsc{MLT-OL}).} Our third method replicates the physics of correct lighting as much as possible to generate photo-realistic rendering. In order to do so, we setup outdoor illumination which is in the form of an environment mapping with real high-definition spherical sky panoramas. The environment map that replicates outdoor lighting is cast through windows and contributes to the indoor lighting naturally. All windows are set as fully transparent to prevent artifacts on glasses and facilitate the outdoor lights to pass through. Person and plant are removed from the scene as the models are not realistic. The default wall texture is set as purely white. We use Mitsuba~\cite{Mitsuba} for physically based rendering. We use Path Space Metropolis Light Transport (MLT) integrator~\cite{veach1997metropolis} since it handles complicate structure and materials more efficiently. A comparison of rendering quality versus time with different integrators is shown in Figure~\ref{fig:rendertech}. We can see that MLT integrator with direct illumination sampler rate 512 produces almost artifact-free renderings with affordable computation time. All the materials are set as two-sided to prevent flipped surface normal.

The images rendered using raw models from SUNCG show severe light leakage in room corners. 
The reason is that the walls, floors, and ceilings are represented by single planar surfaces so light rays can pass through at boundaries.
We fix this problem by assigning walls with thickness (10cm in our experiments) such that each wall is represented by two surfaces. We also force the connecting walls to solidly intersect with each other to prevent light leakage caused by floating number accuracy problems during the rendering.

\mypara{Physically Based Rendering with Indoor Lights (\textsc{MLT-IL/OL}).}  We also setup indoor illumination for light resulting from lighting appliances in the scene. However, the 3D dataset is labeled at the object level (e.g. lamp), and the specific \emph{light generating} parts (e.g. bulb) is unknown. Therefore, we manually labeled all \emph{light generating} parts of objects in order to generate correct indoor lighting. For light appliances that do not have a bulb, representing geometry in cases where bulb is deemed to be not seen, we manually added a spherical bulb geometry at the proper location. The bulb geometries of the lighting appliances are set as area emitter to work as indoor lights. 
% As an efficient approximation of the translucent lamp shade, we set lamp shade geometries to be area emitter as well with very low radiance value. 
Similar to the outdoor lighting, we use Mitsuba and MLT integrator for physically based indoor lights. Figure~\ref{fig:rendering} shows several examples of images generated by different rendering techniques under the same camera. We can see, especially from the zoomed in view, that \textsc{MLT-IL/OL} produces soft shadow and natural looking materials.

\vspace{-1mm}
\subsection{Image Selection}
\begin{figure}
\vspace{-3mm}
\centering
 \includegraphics[width=1\columnwidth]{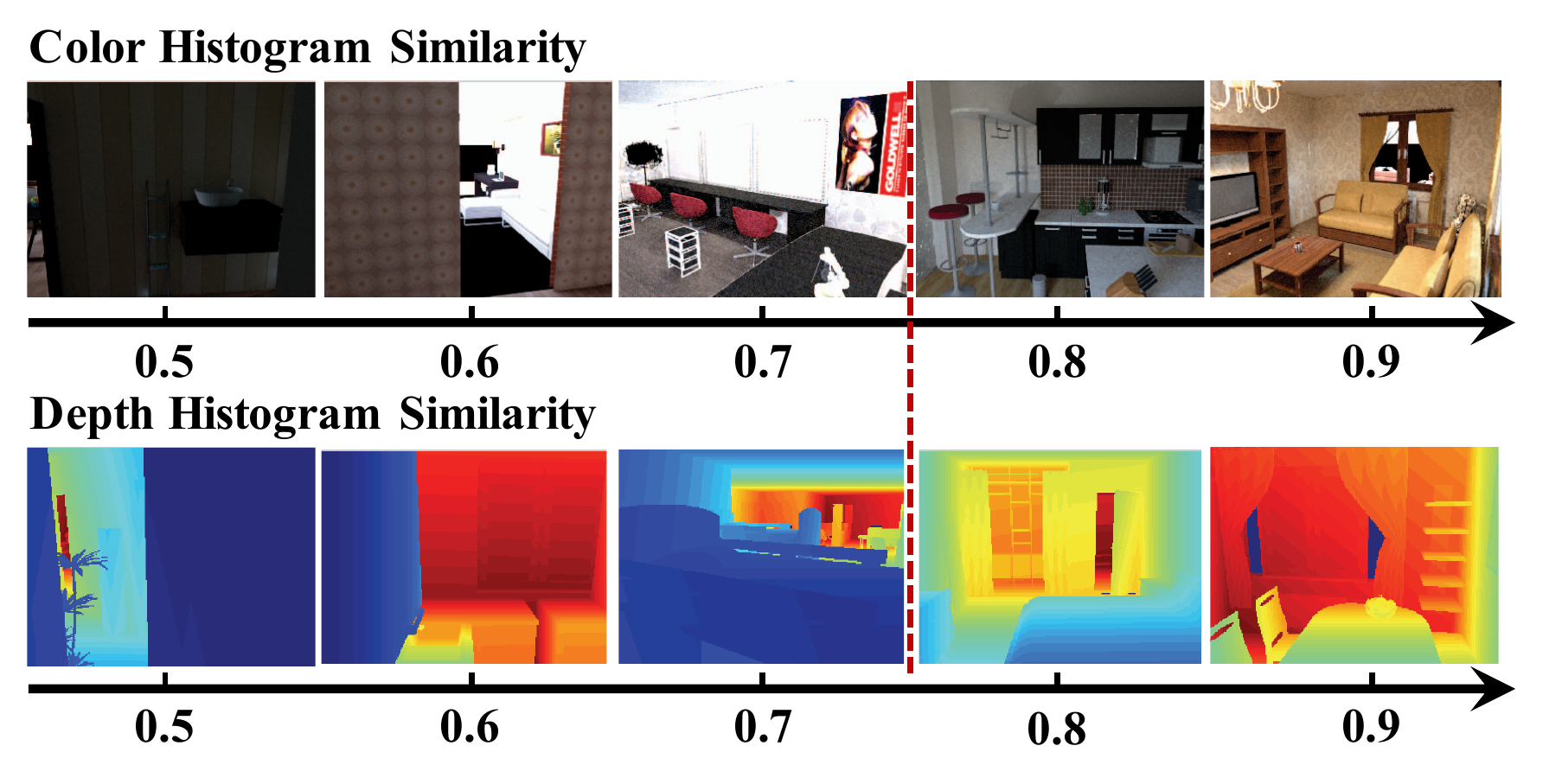}
 \vspace{-5mm}
\caption{\label{fig:histogram} Histogram similarity between synthetic data and real data from NYUv2, based on which we do the image selection.}
\vspace{-4mm}
\end{figure}

\vspace{-1mm}
The final step of our image synthesis pipeline is to select a subset of images to use for training.  Ideally, each of the images in our synthetic training set will be similar to ones found in a test set (e.g., NYUv2). However not all of them are good due to insufficient lighting or atypical distributions of depths (e.g., occlusion by a close-up object). We perform a selection procedure to keep only the images that are similar to those in NYUv2 dataset in terms of color and depth distribution.  Specifically, we first compute a normalized color histogram for each real image in the NYUv2 dataset. For each image rendered by \textsc{MLT-IL/OL}, we also get the normalized color histograms and calculate the histogram similarity with those from NYUv2 as the sum of minimal value of each bin (Figure~\ref{fig:histogram}). Then for each synthesized image, we assign it the largest similarity compared with all NYUv2 images as the score and do the same for the depth channel. Finally, we select all the images with color score and depth score both larger than 0.70. This process selects 568,793 images from the original 779,342 rendered images.  Those images form our synthetic training set, and is referred as \textbf{\textsc{MLT}} in the latter part of this paper.

\vspace{-1mm}
\subsection{Ground Truth Generation}
\vspace{-1mm}
We generate per-pixel ground truth images encoding surface normal, semantic segmentation, and object boundary for each image.  Since we have the full 3D model and camera viewpoints, generating these ground images can be done via rendering with OpenGL (e.g., with an item buffer).

\vspace{-1mm}
\section{Indoor Scene Understanding Tasks}
\vspace{-1mm}
We investigate three fundamental scene understanding tasks: (1) surface normal estimation, (2) semantic segmentation, and (3) object boundary detection. For all tasks we show how our method and synthetic data compares with state of the art works in the literature. Specifically, we compare with Eigen \etal~\cite{multitaskSingleNetwork} for normal estimation, with Long \etal~\cite{long2015fully} and Yu \etal~\cite{dialateSeg} for semantic segmentation, and with Xie \etal~\cite{xie2015holistically} for object boundary detection. We perform these comparisons systematically using different rendering conditions introduced in Section~\ref{sec:data}. In addition, for normal estimation, we also add \emph{object without context} rendering, which allows us to investigate the importance of context when using synthetic data as well.

\vspace{-1mm}
\subsection{Normal Estimation}
\vspace{-1mm}
\paragraph{Method.} We utilize a fully convolutional network~\cite{long2015fully} (FCN) with skip-layers for normal estimation, by combining multi-scale feature maps in VGG-16 network~\cite{simonyan2014very} to perform normal estimation. Specifically, the front-end encoder remains the same as conv1-conv5 in VGG-16, and the decoder is symmetric to the encoder with convolution and unpooling layers. To generate high resolution results and alleviate the vanishing gradient problems, we use skip links between each pair of corresponding convolution layers in downstream and upstream parts of the network. To further compensate the loss of spatial information with max pooling, the network remembers pooling switches in downstream, and uses them as unpooling switches at upstream in the corresponding layer. We use the inverse of the dot product between the ground truth and the estimation as loss function similar to Eigen \etal~\cite{multitaskSingleNetwork}

\mypara{Object without Context.} To facilitate a systematic comparison with object-centric synthetic data, where correct context is missing, we use shapes from ShapeNet\cite{chang2015shapenet}, in addition to the rendering methodologies introduced in Sec.~\ref{renderingmethods}. We randomly pick 3500 models from furniture related categories (e.g. bed, chair, cabinet, etc.) and set up 20 cameras from randomly chosen distances and viewing directions. More specifically, we place the model at the center of a 3D sphere and uniformly sample 162 points on the sphere by subdividing it into faces of an icosahedron. For each camera a random vertex of the icosachedron is selected. This point defines a vector together with the sphere center. The camera is placed at a random distance from the center between $1.5\times$ to $4.5\times$ of object bounding box diagonal, and points towards the center.

\mypara{Training.} We directly pretrain on our synthetic data, followed by finetuning on NYUv2 similar to Bansa \etal~\cite{Bansal16}. 
%We also replicate state of the art training schedules by pretraining on ImageNet, followed directly by finetuning on NYUv2, to verify our implementation and to compare with our results. 
We use RMSprop~\cite{tieleman2012lecture} to train our network. The learning rate is set as $1\times10{-3}$, reducing to half every $300K$ iterations for the pretraining; and $1\times10{-4}$ reducing to half every $10K$ iterations for finetuning. The color image is zero-centered by subtracting $128$. We use the procedure provided by~\cite{silberman2012indoor} to generate the ground truth surface normals on NYUv2 as it provides more local details resulting in more realistic shape representation compared to others~\cite{normalgndsmooth}. The ground truth also provides a score for each pixel indicating if the normal converted from local depth is reliable. We use only reliable pixels during the training.
%Refer to our supplementary material for details on the ground truth generation process.

\begin{table*}
\vspace{-5mm}
\setlength{\tabcolsep}{9 pt}
\centering
\begin{tabular}{ccc|ccccc}
\hline 
Pre-Train & Finetune  & Selection & Mean ($^\circ$) $\downarrow$ & Median($^\circ$) $\downarrow$ & $11.25^\circ$ ($\%$) $\uparrow$ & $22.5^\circ$ ($\%$) $\uparrow$ & $30^\circ$($\%$) $\uparrow$ \tabularnewline
\hline 
\multicolumn{3}{c|}{Eigen \etal~\cite{multitaskSingleNetwork}} & 22.2 & 15.3 & 38.6 & 64.0 & 73.9  \tabularnewline
\multicolumn{3}{c|}{NYUv2} & 27.30 & 21.12 & 27.21 & 52.61 & 64.72 \tabularnewline
\hline 
%\hline
MLT Object & - & - & 48.78 & 47.49 & 3.56 & 12.79 & 21.35 \tabularnewline
%\hline 
\textsc{MLT-OL} & -  & No & 49.33 & 42.30 & 7.47 & 23.24 & 34.09 \tabularnewline
%\hline 
\textsc{MLT-IL/OL} & -  & No & 28.82 & 22.66 & 24.08 & 49.70 & 61.52 \tabularnewline
% \hline 
% OpenGL & N/A & In/Outdoor & Regular &  &  &  &  & \tabularnewline
%\hline 
%MLT & NYUv2 & In/Outdoor & Regular & 23.8934 & 16.6333 & 0.3487 & 0.6167 & 0.7203 \tabularnewline
% \hline 
% OpenGL & NYU & In/Outdoor & Regular &  &  &  &  & \tabularnewline
%\hline 
\textsc{MLT-IL/Ol} & - & Yes & 27.90 & 21.29 & 26.76 & 52.21 & 63.75\tabularnewline
\hline 
%\hline
\textsc{OpenGL-DL} & -  & Yes & 34.02 & 28.00 & 18.56 & 41.14 & 52.90 \tabularnewline
%\hline 
\textsc{OpenGL-IL} & -  & Yes & 33.06 & 26.68 & 20.89 & 43.46 & 54.66  \tabularnewline
\hline 
%\hline
\textsc{OpenGL-IL} & NYUv2 & Yes & 23.38 & 16.12 & 35.98 & 62.93 & 73.17
 \tabularnewline
%\hline 
\textsc{MLT-IL/OL} & NYUv2  & Yes & \textbf{21.74} & \textbf{14.75} & \textbf{39.37} & \textbf{66.25} & \textbf{76.06}
 \tabularnewline
%\hline 

\hline
\end{tabular}
\vspace{1mm}
\protect\caption{Performance of Normal Estimation on NYUv2 with different training protocols. The first three column lists the dataset for pretraining and finetuning, and if image selection is done. The evaluation metrics are mean and median of angular error, and percentage of pixels with error smaller than $11.25^\circ$, $22.5^\circ$, and $30^\circ$.}
\label{tab:normal}
\end{table*}

\begin{figure*}
\centering
 \includegraphics[width=1\textwidth]{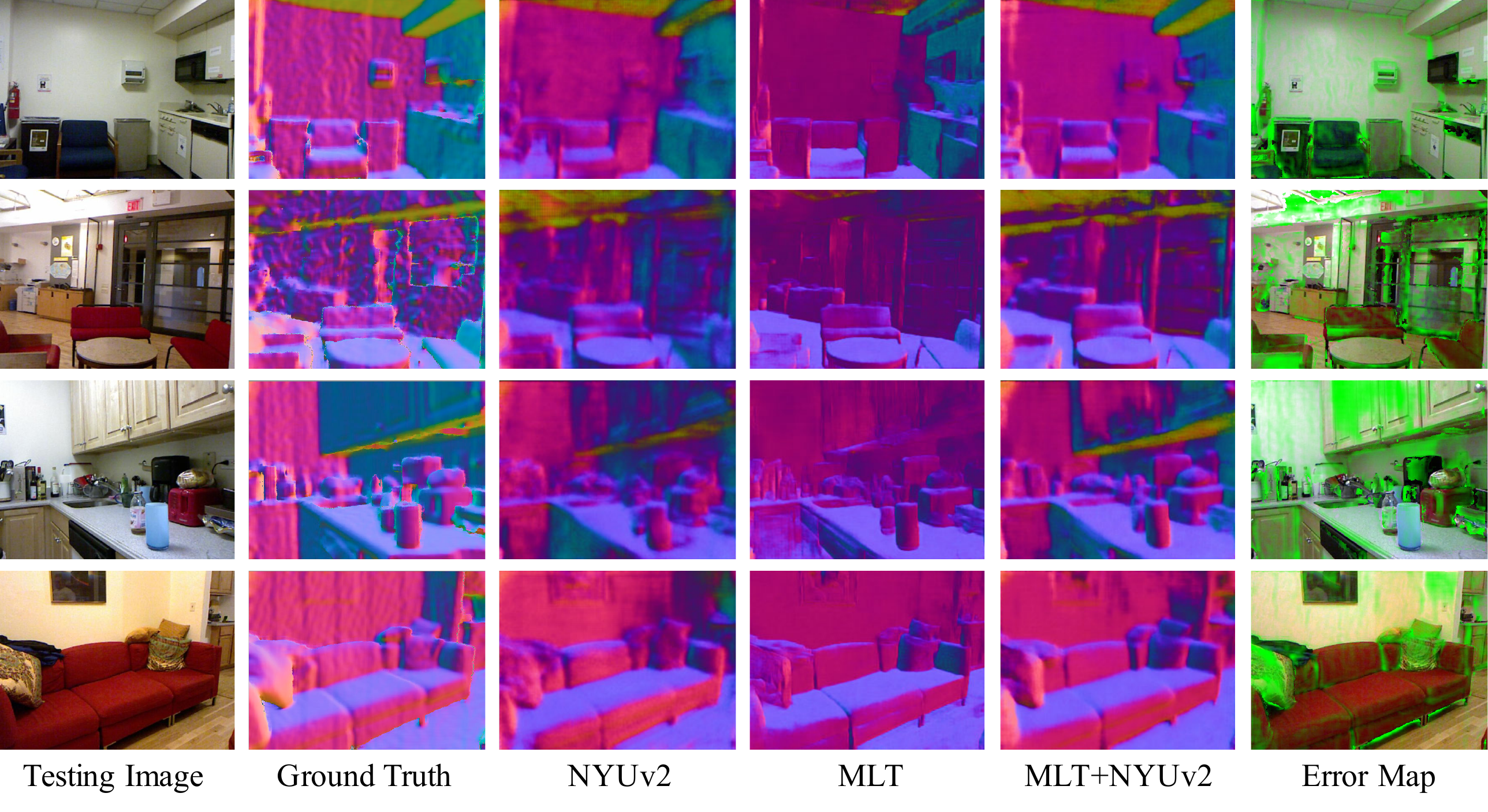}
 \vspace{-6mm}
\caption{\label{fig:normalcompare} {\bf Normal estimation results.} The pretrained model on MLT provides more local details, and model further finetuned on NYUv2 provides the best performance. The last column shows color image overlaid with angular error map. We can see a considerable amount of error happens on wall where ground truth is noisy.}
\vspace{-3mm}
\end{figure*}

\mypara{Experiments.}
We conduct normal estimation experiments on NYUv2 with different training protocols. First, we directly train on NYUv2. Then we pretrain on various of MLT and OpenGL render settings respectively and finetune on NYUv2. Table~\ref{tab:normal} shows the performance. We can see that: 
\vspace{-2mm}
\begin{itemize}
\setlength{\topsep}{0pt}
\setlength{\parsep}{0pt}
\setlength{\parskip}{0pt}
\setlength{\itemsep}{2pt}
    \item The model pretrained on \textsc{MLT} and finetuned on NYUv2 (the last row) achieves the best performance, which outperforms the state of the art.
    \item Without finetuning, pretrained model on \textsc{MLT} significantly outperforms model pretrained on OpenGL based rendering and achieves similar performance with the model directly trained on NYUv2. This shows that physically based rendering with correct illumination is essential to encode useful information for normal prediction task.
    \item The model trained with images after image selection achieves better performance than using all rendered images, which demonstrates that good quality of training image is important for the pretraining.
    \item  The MLT with both indoor and outdoor lighting significantly outperforms the case with only outdoor lighting, which suggests the importance of indoor lighting.
\end{itemize}

Figure~\ref{fig:normalcompare} shows visual results for normal estimation on NYUv2 test split. We can see that the result from the model pretrained on MLT rendering provides sharper edges and more local details compared to the one from the model further finetuned on NYUv2, which is presumably because of the overly smoothed and noisy ground truth. Figure~\ref{fig:normalcompare} last-column visualizes the angular error of our result compared to the ground truth, and we can see that a significant portion of the error concentrates on the walls, where our purely flat prediction is a better representation of wall normals. On the other hand, the ground truth shows significant deviation from the correct normal map. Based on this observation, we highlight the importance of high quality of ground truth. It is clear that training on synthetic data helps our model outperform and correct the NYUv2 ground truth data at certain regions such as large flat areas.

\begin{figure*}[t]
\vspace{-10mm}
\centering
 \includegraphics[width=1\textwidth]{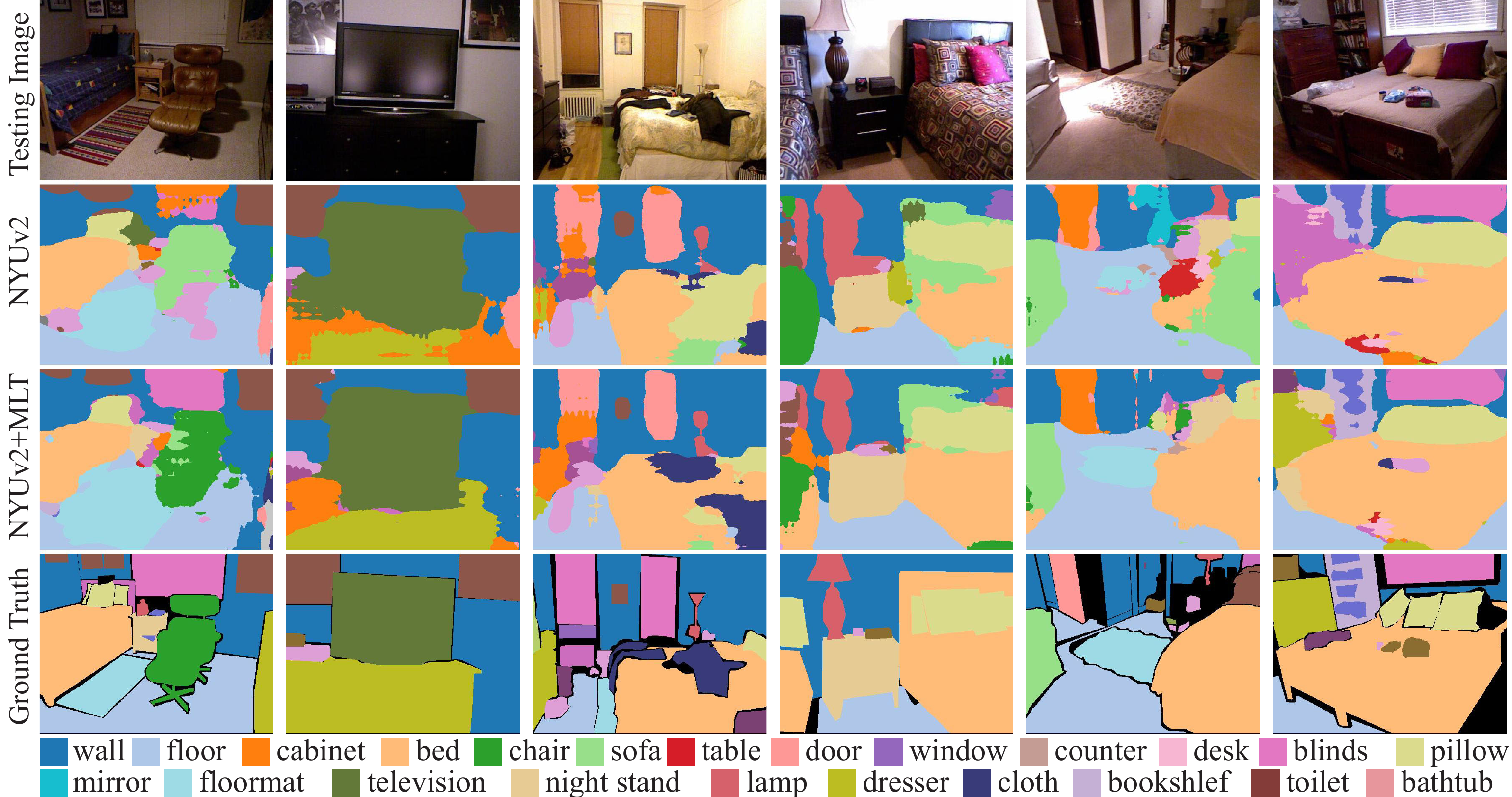}
\caption{\label{fig:segmentation} {\bf Semantic Segmentation results.} The model pretrained on synthetic rendering data gives more accurate segmentation result. For example the model trained only with NYU data mis-labeled the chair, whereas the model pretrained on the synthetic data predicts correctly.}
\vspace{-2mm}
\end{figure*}

\subsection{Semantic Segmentation}
\paragraph{Method.} We use the network model proposed in~\cite{dialateSeg} for semantic segmentation. The network structure is adopted from the VGG-16 network~\cite{simonyan2014very}, however using dilated convolution layers to encode context information, which achieves better performance than~\cite{long2015fully} on NYUv2 in our experiments.
We initialize the weights using the VGG-16 network~\cite{simonyan2014very} trained on ImageNet classification task using the procedure described in~\cite{dialateSeg}. We evaluate on the same 40 semantic classes as~\cite{gupta2013perceptual}. 

\mypara{Training.} To use synthetic data for pretraining, we map our synthetic ground truth labels to the appropriate class name in these 40 classes (note that some categories do not present in our synthetic data).
%\footnote{Refer to our supplementary material for the exact cross-match.}. 
%However, note that there are several objects that are not present in our synthetic data (e.g. books, papers) as shown in Figure~\ref{fig:classdistribution}. 
We first initialize the network with pretrained weights from ImageNet. We then follow with pretraining on our synthetic dataset, and finally finetune on NYUv2. We also replicate the corresponding state of the art training schedules by pretraining on ImageNet, followed directly by finetuning on NYUv2, for comparison. We use stochastic gradient descent with learning rate of $1\times10^{-5}$ for training on synthetic data and NYUv2.

\mypara{Experiments.}
We use the average pixel-level intersection over union (IoU) to evaluate performance on semantic segmentation.
We pretrained the model on our synthetic data with different rendering method: depth, OpenGL color rendering, and MLT color rendering. For the depth based model we encode the depth using HHA same as~\cite{gupta2015aligning}.
Overall, pretraining on synthetic data helps improve the performance in semantic segmentation, compared to directly training on NYUv2 as seen in Figure~\ref{fig:segmentation}, and Table~\ref{tab:seg}. 
This shows that the synthetic data helps the network learn richer high level context information than limited real data.

\begin{figure}
\centering
 \includegraphics[width=1\columnwidth]{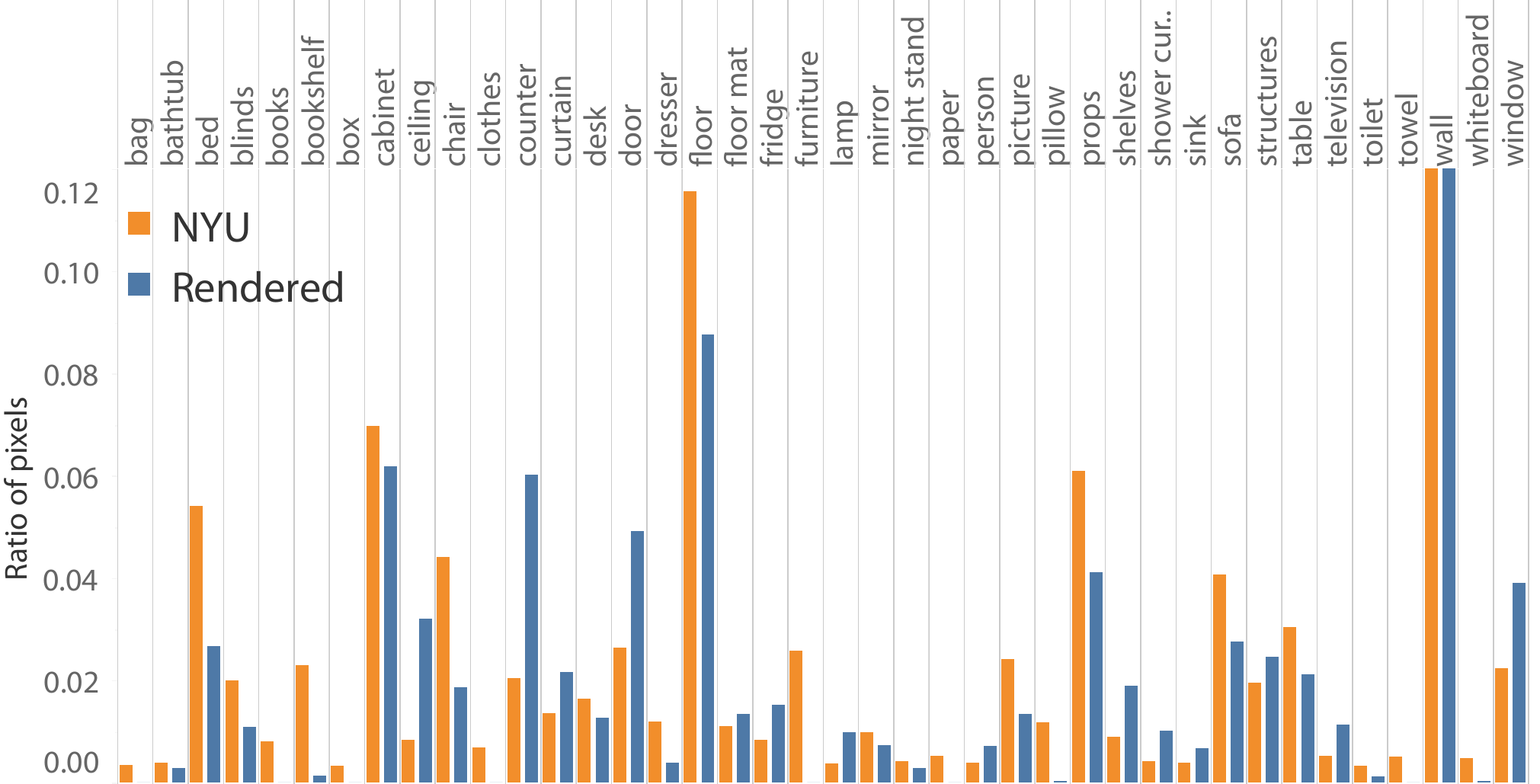}
\caption{\label{fig:classdistribution} Distribution of classes in our data.}
\vspace{-6mm}
\end{figure}

Handa \etal~\cite{handa2015scenenet} use only rendered depth to train their 11 class semantic segmentation model due to the lack of realistic texture and material in their dataset (see HHA results in Table~\ref{tab:seg}). However, our results demonstrate that color information is critical for more fine gained semantic segmentation task: in the 40 class task Model trained with color information achieves significantly better performance. 
For the color based models, pretraining on physically based rendering images helps to achieve better performance than pretraining on  OpenGL rendering. This finding is consistent with normal estimation experiments.
%On the other hand, OpenGL rendering with local lights perform similar to OpenGL rendering without lights.

\begin{table}
\label{tab:seg}
\centering
\setlength{\tabcolsep}{10 pt}
\begin{tabular}{c l c}
\hline 
Input & Pre-train & Mean IoU\tabularnewline
\hline 
\multirow{2}{*}{HHA} & ImageNet  &  27.6\tabularnewline
& ImageNet+OpenGL  &  30.2 \tabularnewline
\hline 
\multirow{5}{*}{RGB} & Long \etal~\cite{long2015fully} & 31.6\tabularnewline
 & Yu \etal~\cite{dialateSeg} & 31.7\tabularnewline
  \cline{2-3} 
 & ImageNet + \textsc{OpenGL} & 32.8\tabularnewline
 & ImageNet + \textsc{MLT} & \textbf{33.2}\tabularnewline
\hline
\end{tabular}
\vspace{1mm}
\protect\caption{Performance of Semantic Segmentation on NYUv2 with different training setting. All models are fine-tuned on NYUv2.}
\vspace{-2mm}
\end{table}

\subsection{Object Boundary Detection}

\paragraph{Method.} We adopt Xie \etal's~\cite{xie2015holistically} network architecture for object boundary detection task as they reported performance on NYUv2. The network starts with the front end of VGG-16, followed by a set of auxiliary-output layers, which produce boundary maps in multiple scales from fine to coarse. 
%Each of these boundary maps are trained under supervision of the ground truth on corresponding scale. 
A weighted-fusion layer then learns the weights to combine boundary outputs in multi-scale to produce the final result.
%(refer to our supplementary material for the detailed network structure).
To evaluate the network, we follow the setting in~\cite{gupta2013perceptual}, where the boundary ground truth is defined as the boundary of instance level segmentation.

\begin{figure*}
\vspace{-10mm}
\centering
\includegraphics[width=1\textwidth]{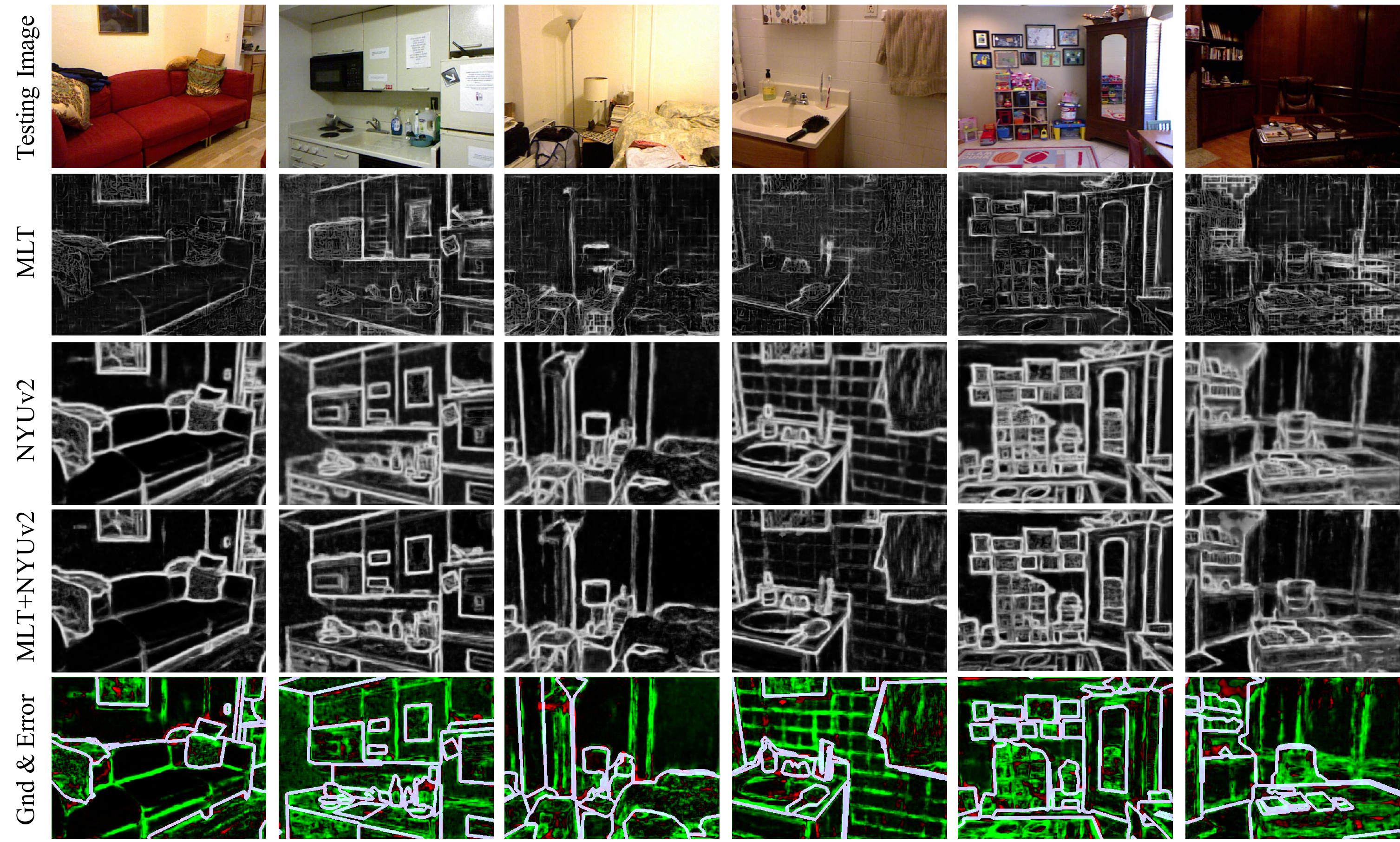}
\caption{\label{fig:boundarycompare} {\bf Boundary estimation results.} The last row shows ground truth overlaid with the difference between model without (NYUv2) and with (MLT+NYUv2) synthetic data pretraining. Red and green indicates pixels enhanced and suppressed by MLT+NYUv2. The model with synthetic data pretraining successfully suppresses texture and background edges compared to the model without.}
\vspace{-4mm}
\end{figure*}

\mypara{Training.} Similar to the semantic segmentation, we first initialize the network with pretrained weights on ImageNet. We then pretrain on our synthetic dataset, and finetune on NYUv2. 
We also replicate the state of the art training procedure by pretraining on ImageNet, and directly finetune on NYUv2, for comparison. To highlight the difference between multiple rendering techniques, we only train on color image without using depth.
We follow the same procedure introduced in~\cite{xie2015holistically}. The standard stochastic gradient descend is used for optimization. The learning rate is initially set to be smaller ($2\times10^{-7}$) to deal with larger image resolution of NYUv2, and is reduced even more, to $1/10$ after each $10K$ iterations on NYUv2. For synthetic data, similar to our procedure in like normal estimation task, the learning rate is reduced every $300k$ iterations.

\begin{table}
\centering
\setlength{\tabcolsep}{5 pt}
\begin{tabular}{cc|cccc}
\hline 
Pre-train & Finetune & OSD$\uparrow$ & OIS$\uparrow$ & AP$\uparrow$ & R50$\uparrow$\tabularnewline
\hline 
%\hline 
NYUv2\cite{xie2015holistically} & - & 0.713 & 0.725 & 0.711 & 0.267 \tabularnewline
%\hline 
\textsc{OpenGL-IL} & - & 0.523 & 0.555 & 0.511 & 0.504 \tabularnewline
%\hline 
\textsc{MLT-IL/OL} & - & 0.604 & 0.621 & 0.587 & 0.749 \tabularnewline
%\hline 
\textsc{OpenGL-IL} & NYUv2 & 0.716 & 0.729 & 0.715 & \textbf{0.893} \tabularnewline
%\hline 
\textsc{MLT-IL/OL} & NYUv2 & \textbf{0.725} & \textbf{0.736} & \textbf{0.720} & 0.887 \tabularnewline
\hline 

\end{tabular}

\protect\caption{Performance of boundary detection on NYUv2}
\label{tab:boundaryresult}
\vspace{-6mm}
\end{table}

\vspace{-1mm}
\mypara{Experiments.}
We train the model proposed in Xie \etal's~\cite{xie2015holistically} with multiple different protocols and show our comparison and evaluation on NYUv2 in Table~\ref{tab:boundaryresult}. Following the setting of~\cite{xie2015holistically}, we take the average of the output from 2nd to 4th multiscale layers as the final result and perform non-maximum suppression and edge thinning. We use the ground truth in~\cite{gupta2013perceptual}, and evaluation metrics in~\cite{dollar2015fast}. 

We train with the code released by~\cite{xie2015holistically} and achieve the performance shown in the first row of Table~\ref{tab:boundaryresult}. We could not replicate the exact number in the paper but we were fairly close, which might be due to the randomized nature of training procedure. We first finetune the model based on the ImageNet initialization on the synthetic dataset and further finetune on NYUv2. Table~\ref{tab:boundaryresult} shows that the synthetic data pretraining provides consistent improvement on all evaluation metrics. Consistently, we see the model pretrained with MLT rendering achieves the best performance. 

Figure~\ref{fig:boundarycompare} shows a comparison between results from different models. Pretrained model on synthetic data, prior to finetuning on real data produces sharper results but is more sensitive to noise. The last column highlights the difference between model with and without pretraining on our synthetic data. We can see that edges within objects themselves as well as the ones in the background (green) are suppressed and true object boundary (red) are enhanced by the model with pretraining on synthetic.

\vspace{-1mm}
\section{Conclusion}
\vspace{-1mm}
We introduce a large-scale synthetic dataset with 500K rendered images of contextually meaningful 3D indoor scenes with different lighting and rendering settings, as well as indoor scenes models they were rendered from. 
We show that pretraining on our physically based rendering with realistic lighting boosts the performance of indoor scene understanding tasks upon the state of the art methods.
%, and 
%with realistic indoor and outdoor lights 
%boost indoor scene understanding tasks' performance. Our experiments show that our approach 
%enable us to perform better than state of the art in surface normal estimation, semantic segmentation, and edge detection tasks.

% \begin{figure}
% \centering
%  \includegraphics[width=1\columnwidth]{figures/comp_dummy.png}
% \caption{\label{fig:normalviserror} Visualization of error distribution on NYUv2 testing images.}
% \end{figure} 

\vspace{-2mm}
\section*{Acknowledgments}
\vspace{-2mm}
This work is supported by Adobe, Intel, Facebook, and NSF 
(IIS-1251217 and VEC 1539014/ 1539099). 
It makes use of data from Planner5D and hardware
provided by NVIDIA and Intel.

\clearpage
{\small
\bibliographystyle{ieee}
\bibliography{egbib}

\begin{thebibliography}{10}\itemsep=-1pt

\bibitem{Mitsuba}
{Mitsuba} physically based renderer.
\newblock \url{http://www.mitsuba-renderer.org/}.

\bibitem{aubry2015understanding}
M.~Aubry and B.~C. Russell.
\newblock Understanding deep features with computer-generated imagery.
\newblock In {\em Proceedings of the IEEE International Conference on Computer
  Vision}, pages 2875--2883, 2015.

\bibitem{Bansal16}
A.~Bansal, B.~C. Russell, and A.~Gupta.
\newblock Marr revisited: {2D-3D} alignment via surface normal prediction.
\newblock In {\em Conference on Computer Vision and Pattern Recognition}, 2016.

\bibitem{chang2015shapenet}
A.~X. Chang, T.~Funkhouser, L.~Guibas, P.~Hanrahan, Q.~Huang, Z.~Li,
  S.~Savarese, M.~Savva, S.~Song, H.~Su, et~al.
\newblock Shapenet: An information-rich 3d model repository.
\newblock {\em arXiv preprint arXiv:1512.03012}, 2015.

\bibitem{dollar2015fast}
P.~Doll{\'a}r and C.~L. Zitnick.
\newblock Fast edge detection using structured forests.
\newblock {\em IEEE transactions on pattern analysis and machine intelligence},
  37(8):1558--1570, 2015.

\bibitem{dosovitskiy2015flownet}
A.~Dosovitskiy, P.~Fischery, E.~Ilg, C.~Hazirbas, V.~Golkov, P.~van~der Smagt,
  D.~Cremers, T.~Brox, et~al.
\newblock Flownet: Learning optical flow with convolutional networks.
\newblock In {\em 2015 IEEE International Conference on Computer Vision
  (ICCV)}, pages 2758--2766. IEEE, 2015.

\bibitem{multitaskSingleNetwork}
D.~Eigen and R.~Fergus.
\newblock Predicting depth, surface normals and semantic labels with a common
  multi-scale convolutional architecture.
\newblock In {\em Proceedings of the IEEE International Conference on Computer
  Vision}, pages 2650--2658, 2015.

\bibitem{everingham2010pascal}
M.~Everingham, L.~Van~Gool, C.~K. Williams, J.~Winn, and A.~Zisserman.
\newblock The pascal visual object classes (voc) challenge.
\newblock {\em International journal of computer vision}, 88(2):303--338, 2010.

\bibitem{gupta2015aligning}
S.~Gupta, P.~Arbel{\'a}ez, R.~Girshick, and J.~Malik.
\newblock Aligning 3d models to rgb-d images of cluttered scenes.
\newblock In {\em Proceedings of the IEEE Conference on Computer Vision and
  Pattern Recognition}, pages 4731--4740, 2015.

\bibitem{gupta2013perceptual}
S.~Gupta, P.~Arbelaez, and J.~Malik.
\newblock Perceptual organization and recognition of indoor scenes from rgb-d
  images.
\newblock In {\em Proceedings of the IEEE Conference on Computer Vision and
  Pattern Recognition}, pages 564--571, 2013.

\bibitem{handa2015scenenet}
A.~Handa, V.~Patraucean, V.~Badrinarayanan, S.~Stent, and R.~Cipolla.
\newblock Scenenet: Understanding real world indoor scenes with synthetic data.
\newblock {\em arXiv preprint arXiv:1511.07041}, 2015.

\bibitem{handa:etal:ICRA2014}
A.~Handa, T.~Whelan, J.~McDonald, and A.~Davison.
\newblock A benchmark for {RGB-D} visual odometry, {3D} reconstruction and
  {SLAM}.
\newblock In {\em IEEE Intl. Conf. on Robotics and Automation, ICRA}, Hong
  Kong, China, May 2014.

\bibitem{EvalVirtualWorld}
B.~Kaneva, A.~Torralba, and W.~T. Freeman.
\newblock Evaluation of image features using a photorealistic virtual world.
\newblock In {\em 2011 International Conference on Computer Vision}, pages
  2282--2289. IEEE, 2011.

\bibitem{MSCOCO}
T.-Y. Lin, M.~Maire, S.~Belongie, J.~Hays, P.~Perona, D.~Ramanan,
  P.~Doll{\'a}r, and C.~L. Zitnick.
\newblock Microsoft coco: Common objects in context.
\newblock In {\em European Conference on Computer Vision}, pages 740--755.
  Springer, 2014.

\bibitem{long2015fully}
J.~Long, E.~Shelhamer, and T.~Darrell.
\newblock Fully convolutional networks for semantic segmentation.
\newblock In {\em Proceedings of the IEEE Conference on Computer Vision and
  Pattern Recognition}, pages 3431--3440, 2015.

\bibitem{normalgndsmooth}
B.~Z. L’ubor~Ladick{\`y} and M.~Pollefeys.
\newblock Discriminatively trained dense surface normal estimation.

\bibitem{HowUseful}
Y.~Movshovitz-Attias, T.~Kanade, and Y.~Sheikh.
\newblock How useful is photo-realistic rendering for visual learning?
\newblock {\em arXiv preprint arXiv:1603.08152}, 2016.

\bibitem{richter2016playing}
S.~R. Richter, V.~Vineet, S.~Roth, and V.~Koltun.
\newblock Playing for data: Ground truth from computer games.
\newblock In {\em European Conference on Computer Vision}, pages 102--118.
  Springer, 2016.

\bibitem{ros2016synthia}
G.~Ros, L.~Sellart, J.~Materzynska, D.~Vazquez, and A.~M. Lopez.
\newblock The synthia dataset: A large collection of synthetic images for
  semantic segmentation of urban scenes.
\newblock In {\em Proceedings of the IEEE Conference on Computer Vision and
  Pattern Recognition}, pages 3234--3243, 2016.

\bibitem{SSCNet}
S.~Shuran, Y.~Fisher, Z.~Andy, X.~C. Angel, S.~Manolis, and F.~Thomas.
\newblock {S}emantic {S}cene {C}ompletion from a {S}ingle {D}epth {I}mage.
\newblock In {\em arXiv}, 2016.

\bibitem{silberman2012indoor}
N.~Silberman, D.~Hoiem, P.~Kohli, and R.~Fergus.
\newblock Indoor segmentation and support inference from rgbd images.
\newblock In {\em European Conference on Computer Vision}, pages 746--760.
  Springer, 2012.

\bibitem{simonyan2014very}
K.~Simonyan and A.~Zisserman.
\newblock Very deep convolutional networks for large-scale image recognition.
\newblock {\em arXiv preprint arXiv:1409.1556}, 2014.

\bibitem{SUNRGBD}
S.~Song, S.~Lichtenberg, and J.~Xiao.
\newblock {SUN RGB-D}: A {RGB-D} scene understanding benchmark suite.
\newblock In {\em CVPR}, 2015.

\bibitem{su2015render}
H.~Su, C.~R. Qi, Y.~Li, and L.~J. Guibas.
\newblock Render for cnn: Viewpoint estimation in images using cnns trained
  with rendered 3d model views.
\newblock In {\em Proceedings of the IEEE International Conference on Computer
  Vision}, pages 2686--2694, 2015.

\bibitem{tieleman2012lecture}
T.~Tieleman and G.~Hinton.
\newblock Lecture 6.5-rmsprop: Divide the gradient by a running average of its
  recent magnitude.
\newblock {\em COURSERA: Neural Networks for Machine Learning}, 4(2), 2012.

\bibitem{veach1997metropolis}
E.~Veach and L.~J. Guibas.
\newblock Metropolis light transport.
\newblock In {\em Proceedings of the 24th annual conference on Computer
  graphics and interactive techniques}, pages 65--76. ACM Press/Addison-Wesley
  Publishing Co., 1997.

\bibitem{xiang2016objectnet3d}
Y.~Xiang, W.~Kim, W.~Chen, J.~Ji, C.~Choy, H.~Su, R.~Mottaghi, L.~Guibas, and
  S.~Savarese.
\newblock Objectnet3d: A large scale database for 3d object recognition.
\newblock In {\em European Conference on Computer Vision}, pages 160--176.
  Springer, 2016.

\bibitem{xie2015holistically}
S.~Xie and Z.~Tu.
\newblock Holistically-nested edge detection.
\newblock In {\em Proceedings of the IEEE International Conference on Computer
  Vision}, pages 1395--1403, 2015.

\bibitem{dialateSeg}
F.~Yu and V.~Koltun.
\newblock Multi-scale context aggregation by dilated convolutions.
\newblock In {\em ICLR}, 2016.

\bibitem{zhang2016deep}
Y.~Zhang, M.~Bai, P.~Kohli, S.~Izadi, and J.~Xiao.
\newblock Deepcontext: Context-encoding neural pathways for 3d holistic scene
  understanding.
\newblock {\em arXiv preprint arXiv:1603.04922}, 2016.

\end{thebibliography}
}

\end{document}